\documentclass[10pt,twocolumn,letterpaper]{article}

\usepackage[pagenumbers]{cvpr} 

\newcommand{\customfootnotetext}[2]{{
\renewcommand{\thefootnote}{#1}
\footnotetext[0]{#2}}}

\definecolor{cvprblue}{rgb}{0.21,0.49,0.74}
\usepackage[pagebackref,breaklinks,colorlinks,allcolors=cvprblue]{hyperref}
\usepackage{algorithm}
\usepackage{algorithmic}
\usepackage{amsmath,amssymb}
\usepackage[accsupp]{axessibility}
\usepackage{makecell}
\usepackage{booktabs}
\usepackage{url}
\usepackage {colortbl,array,xcolor}
\usepackage{multirow}
\usepackage{comment}
\usepackage{arydshln}
\usepackage{cite}
\usepackage{siunitx}
\usepackage{graphicx}
\usepackage{pifont}
\usepackage{bm}

\newcommand{\vI}{\bm{I}}
\definecolor{Gray}{gray}{0.95}
\newcolumntype{g}{>{\columncolor{Gray}}c}

\title{Learning from Synthetic Data via Provenance-Based Input Gradient Guidance}

\author{Koshiro Nagano$^1$
\quad
Ryo Fujii$^1$
\quad
Ryo Hachiuma$^2$
\quad
Fumiaki Sato$^3$
\quad
Taiki Sekii$^{3,\dagger}$
\quad
Hideo Saito$^1$
\\
$^{1}$Keio University
\quad
$^{2}$Independent Researcher
\quad
$^{3}$CyberAgent
\\
{\tt\small \{koshiro.nagano, taiki.sekii\}@gmail.com} \\
}

\begin{document}
\twocolumn[{%
\renewcommand\twocolumn[1][]{#1}%
\maketitle
\centering
\vspace{-2mm}
\includegraphics[width=0.9\linewidth]{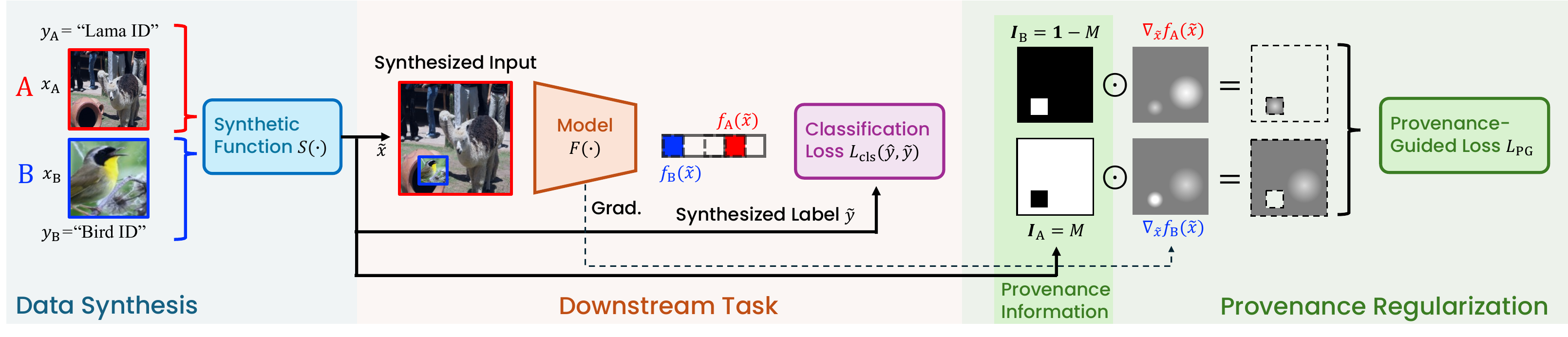}
\vspace{-1mm}
\captionof{figure}{Overview of the proposed method. We adopt CutMix as the synthesis function $S(\cdot)$ and suppress input gradients using the provenance mask $M$, which is automatically obtained during input data synthesis, for auxiliary supervision.
See the main text for details.
}
\label{fig:method}
\vspace{1.5em} 
}]
\customfootnotetext{}{$\dagger$ Corresponding author. }
\begin{abstract}
Learning methods using synthetic data have attracted attention as an effective approach for increasing the diversity of training data while reducing collection costs, thereby improving the robustness of model discrimination.
However, many existing methods improve robustness only indirectly through the diversification of training samples and do not explicitly teach the model which regions in the input space truly contribute to discrimination; consequently, the model may learn spurious correlations caused by synthesis biases and artifacts.
Motivated by this limitation, this paper proposes a learning framework that uses provenance information obtained during the training data synthesis process, indicating whether each region in the input space originates from the target object, as an auxiliary supervisory signal to promote the acquisition of representations focused on target regions.
Specifically, input gradients are decomposed based on information about target and non-target regions during synthesis, and input gradient guidance is introduced to suppress gradients over non-target regions.
This suppresses the model’s reliance on non-target regions and directly promotes the learning of discriminative representations for target regions. Experiments demonstrate the effectiveness and generality of the proposed method across multiple tasks and modalities, including weakly supervised object localization, spatio-temporal action localization, and image classification.
\end{abstract}

\section{Introduction}\label{sec:intro}
Deep learning has become a key technology not only for computer vision tasks such as image classification and object detection~\citep{Redmon_2016_CVPR, Ren_2015_NeurIPS}, but also across many other domains, including natural language processing~\citep{Devlin2019BERT, Brown2020GPT3} and speech recognition~\citep{Baevski2020wav2vec2,Amodei2016DeepSpeech2}. Beyond advances in deep neural network (DNN) architectures, this progress has been driven by improvements in computational environments for training, which have enabled pretraining on large-scale datasets. Such large-scale training endows DNN models with a level of generalization that was previously unattainable, making it possible to acquire robust representations that are less dependent on any specific dataset.
In practice, however, constructing datasets that sufficiently cover the complex and diverse conditions encountered in real-world applications remains challenging owing to the human cost of data collection and annotation and the financial cost of hardware procurement. When training data diversity is limited, a gap arises between the distribution of input data during training and that encountered during deployment. This distribution gap is caused, for example, by changes in background, illumination, and viewpoint, as well as co-occurrence relationships among objects, and it degrades the generalizability of the representations learned by the model. This problem is particularly serious in applications demanding high levels of generalizability and robustness, such as robotics and autonomous agents.

\subsection{Limitations of Prior Work}
As one approach to this problem, learning methods that use synthetic data\footnote{In this paper, we regard samples generated by software simulators, data augmentation, or generative models as synthetic data.} (hereafter referred to as synthetic learning methods) have been actively studied. By using synthetic data, it becomes possible not only to collect the data required to achieve the desired generalization at lower cost, but also to learn from samples that are difficult to collect in the real world, including variations in object categories, appearance, and pose, as well as background and object co-occurrence.
One such line of work comprises synthetic learning methods based on data mixing, such as mixup~\citep{Zhang_2018_ICLR} and CutMix~\citep{Yun_2019_ICCV}. These methods synthesize challenging training samples to recognize by combining multiple images and encourage regularization through linear interpolation of supervisory labels. 
Methods have also been proposed to augment existing training samples using generative models such as generative adversarial networks and diffusion models~\citep{dunlap2023alia}. 
By varying the text used to condition generation, they can introduce objects and backgrounds from classes absent in the collected training data, thereby improving the classifier's robustness.

Despite such progress, most conventional synthetic learning methods improve model robustness only indirectly by diversifying the distribution of training samples in the input space, without explicitly instructing the model which input regions truly contribute to classification~\citep{geirhos2020shortcut}. Specifically, prior studies employ strategies such as mixing images, perturbing textures, and changing backgrounds or context, presenting misleading cues, such as out-of-distribution backgrounds or object co-occurrence relations that induce incorrect classification, through positive and negative examples. 
Consequently, the robustness acquired by the model remains a side effect of training sample augmentation rather than the result of directly learning features effective for recognizing the target object. 
In contrast, this can cause the model to mistakenly learn synthetic biases and artifacts, \ie, distributions introduced by augmentation that differ from the real distribution, preventing model accuracy from scaling with data volume.
In other words, prior work relies solely on supervisory labels: the model must determine by itself, during training, which input regions are specific to the target object, without receiving explicit instruction about the true target regions (\eg, pixel regions or patches containing the target object), despite the fact that, in principle, the synthetic process can identify provenance information indicating which pixels originate from which target object.

\subsection{Overview and Contributions}
Building on this observation, this paper tackles the problem of guiding the model to directly learn discriminative representations of regions corresponding to the target object in the input space (hereafter referred to as target regions), by treating provenance information obtained during the synthetic process as an auxiliary supervisory signal. Specifically, we propose a new learning framework for synthetic data that uses target region information from the synthetic process to instruct the model which input regions should be learned and which should be ignored.
This instruction is realized by input gradient guidance, which separates gradients\footnote{Hereafter, for simplicity, we refer to the gradients obtained by differentiating the model output or loss function with respect to elements in the input space (\eg, pixel values) simply as input gradients.} in the input space between target and non-target regions and optimizes a penalty term, called the “provenance loss,” that suppresses gradients over non-target regions.
Compared with prior work, the proposed method has two key properties through input gradient guidance: (1) it suppresses the model output (\eg, logits) from being driven by non-target regions, and (2) it directly acquires target-focused representations rather than relying on indirect regularization through training sample augmentation. Moreover, the proposed method is modality-agnostic and independent of any specific synthetic learning method. Provided that the synthetic process can identify non-target regions in the input space that cause spurious correlations, the proposed method can be introduced without annotation cost to a wide range of synthetic learning methods, from simple mixing methods such as CutMix to methods based on image generation models.
In experiments, the proposed method is compared with recent state-of-the-art (SoTA) methods across multiple tasks, including weakly supervised object localization, weakly supervised spatio-temporal action localization, and image classification, and its effectiveness is validated through comprehensive ablation studies.

The contributions of this work are twofold: (1) we show that input gradient guidance based on provenance information from the synthetic process promotes the learning of target region-focused representations and suppresses spurious correlations, and (2) we demonstrate that this finding generalizes across multiple tasks and modalities.

\section{Related Work}

\subsection{Suppressing Spurious Correlations}
Spurious correlations refer to cases where a model mistakenly learns apparent correlations or biases unrelated to the target object, such as background or co-occurrence relationships among objects, instead of learning the true characteristics of the target object. For example, when a particular background is frequently observed together with the target object, the model may incorrectly associate the background with target-object features~\citep{Ribeiro_2016_KDD}.
Such degradation in recognition robustness becomes particularly pronounced under domain shift~\citep{geirhos2020shortcut}.
It has also been noted that, in weakly supervised learning for video recognition~\citep{Shao_2026_TMM}, learning tends to rely on object parts or background rather than the target action.
Various methods have been proposed to suppress this problem. These include approaches that adjust the final layer to reduce the contribution of non-robust features~\citep{evals_llr}, methods that identify and remove features unrelated to the target object~\citep{feature_sieve}, and attention-based learning~\citep{Fukui_2019_CVPR} that guides the model toward regions consistent with human-interpretable visual evidence. However, these methods are limited in applicability because they require manual annotations, such as auxiliary labels related to attributes, capture environments, and gaze regions, and impose architectural constraints on DNNs.

Similar to the proposed method, prior work \citep{rrr} operates on input gradients, regularizing them to align with human gaze regions. The proposed method nonetheless differs in two key aspects: (1) it generalizes the provenance loss to the multi-class setting, and (2) it proposes a learning framework that introduces input gradient guidance into synthetic-data learning.

\subsection{Learning with Synthetic Data}\label{sec:synthetic}
Synthetic learning methods that suppress spurious correlations through training data augmentation have been actively studied in recent years. Prior work can be broadly categorized into three approaches: using simulators, data augmentation, and generative models, each of which is described below.

\subsubsection{Use of Simulators}
Synthetic learning methods based on simulators have been widely studied for constructing large-scale datasets under a variety of software-controlled conditions. Representative examples include studies that synthesized urban driving-environment data using GTA-V~\citep{Richter2016GTA} and studies that reproduced diverse human actions and poses~\citep{Varol2017SURREAL}. While these studies can automatically obtain detailed annotations, a key challenge is the human cost of developing the simulator. 
Furthermore, because synthetic data exhibit a domain gap relative to natural images, domain adaptation is often required in practice~\citep{Tobin_2017_IROS}.

\subsubsection{Data Augmentation}
To avoid both the simulator development cost and the domain gap problem, many studies have explored synthesizing unseen data from real images. In addition to classical methods based on geometric and photometric transformations~\citep{Krizhevsky_2012_NeurIPS, He_2016_CVPR}, methods have been proposed that improve robustness by randomly masking divided image patches~\citep{Singh_2017_ICCV} and by mixing multiple images and labels~\citep{Zhang_2018_ICLR, Yun_2019_ICCV}. A fundamental limitation of these approaches, however, is that they cannot synthesize data with diversity beyond the distribution of the original training samples, such as variations in capture environments or target-object appearance.

\subsubsection{Use of Image Generation Models}
Image generation models have greatly advanced synthetic learning methods in recent years~\citep{dunlap2023alia,sariyildiz2023fake,fan2024scaling}. In particular, diffusion models capable of generating high-fidelity images can precisely control attributes such as the types and appearances of background and foreground objects and camera position. For example, Stable Diffusion~\citep{rombach2022high} can edit specific regions in a real image via text prompts, enabling parts of a real image to be transformed into photorealistic elements that do not actually exist. Based on this capability, learning methods have been proposed that generate task-relevant images with diverse appearances and incorporate them into training~\citep{dunlap2023alia}. Compared with simulators or data augmentation, these methods can express synthetic regions more naturally, reproducing data distributions closer to real environments and thereby improving model robustness. However, because image generation models can still produce synthetic artifacts, model accuracy does not necessarily scale with data volume~\citep{dunlap2023alia}.

The prior work described above focuses on diversifying training samples, and model recognition accuracy improves only indirectly as a side effect of increased training data. In contrast, the proposed method uses target region information from the synthetic process as a supervisory signal to directly suppress spurious correlations, including backgrounds and object co-occurrences unrelated to the target object in real images, as well as biases and artifacts introduced during synthesis.

\section{Proposed Method}\label{method}

\subsection{Overview}\label{subsec:overview}
As shown in \cref{fig:method}, the proposed method is a learning framework composed of three elements: (1) synthesis of training data, (2) learning for the downstream task, and (3) regularization using provenance information. As in prior work on synthetic learning methods, the synthesized training samples and supervisory labels are used for downstream-task learning. Regularization is additionally introduced, using the provenance information obtained during synthesis as an auxiliary supervisory signal, to suppress the model from relying on input regions unrelated to the target object.

The proposed method uses provenance information through the following two processes.
\begin{itemize}

\item Provenance extraction: The synthesis function $S(\cdot)$ is applied to input data $x$ (\eg, an image or a set of skeleton points) to generate synthetic data $\tilde{x}$, supervisory labels, and provenance information $\vI$, which indicates the supervisory label to which each element of $\tilde{x}$ (\eg, a pixel or skeleton point) belongs.

\item Input gradient guidance: Spurious correlations are suppressed by introducing the loss $L_{\mathrm{PG}}$, which regularizes the input gradients to be consistent with the provenance information (referred to as the provenance loss).

\end{itemize}

The model is trained using both the downstream-task loss $L_{\mathrm{cls}}$ for the classification problem and the provenance loss $L_{\mathrm{PG}}$.
$L_{\mathrm{cls}}$ is computed as the cross-entropy between the model output $\hat{y}$ for the synthetic data $\tilde{x}$ and the synthetic label $\tilde{y}$.
The total loss function during training is defined as follows:
\begin{equation}
L_{\mathrm{total}} = L_{\mathrm{cls}} + \alpha L_{\mathrm{PG}},
\label{eq:overview_total}
\end{equation}
where $\alpha$ is a coefficient controlling the strength of regularization by input gradient guidance.

\begin{figure*}[t]
\begin{center}
\includegraphics[width=\linewidth]{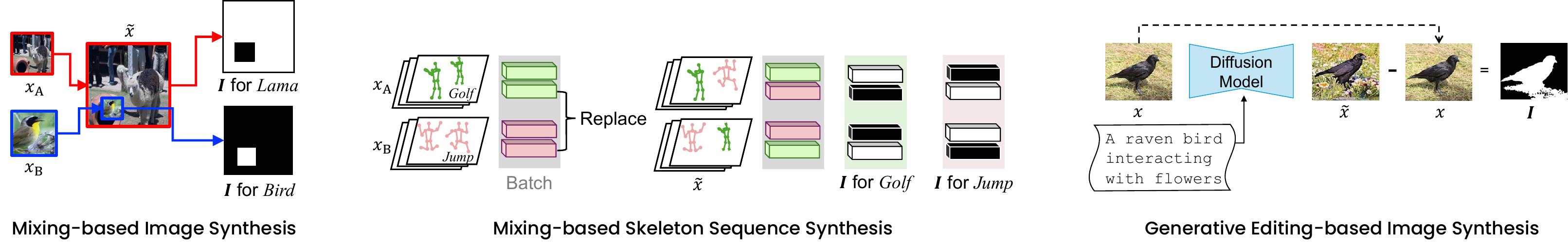}
\caption{Examples of provenance information $\vI$ obtained during synthesis. $\vI$ corresponds to each supervisory label.}
\vspace{-2.0em}
\label{fig:extract}
\end{center}
\end{figure*}

\subsection{Provenance extraction}\label{subsec:eskm}
In this paper, we consider three types of synthesis methods $S(\cdot)$ and describe how the synthetic data $\tilde{x}$ and the provenance information $\vI$ are computed.
The overall process of data synthesis and provenance extraction is illustrated in \cref{fig:extract}.

\subsubsection{Image mixing}\label{subsubsec:image}
Synthetic learning methods that mix images~\citep{Yun_2019_ICCV,qin2020resizemix,kim2020puzzle} create new synthetic data $(\tilde{x}, \tilde{y})$ by combining two samples, $(x_\mathrm{A}, y_\mathrm{A})$ and $(x_\mathrm{B}, y_\mathrm{B})$.
Specifically, a binary mask image $M \in \{0,1\}^{H \times W}$ is first created, and for each location in the synthetic image $\tilde{x}$, it determines whether the value is taken from $x_\mathrm{A}$ ($M(u,v)=1$) or $x_\mathrm{B}$ ($M(u,v)=0$).
For example, in CutMix~\citep{Yun_2019_ICCV}, $M$ contains a rectangular region.
$\tilde{x}$ is computed as follows:
\begin{equation}
\tilde{x} = M \odot x_\mathrm{A} + (\mathbf{1}-M) \odot x_\mathrm{B},
\label{eq:img_mix_synth}
\end{equation}
where $\odot$ denotes the element-wise product.
The supervisory label is computed as a soft label $\tilde{y}$ using the mixing ratio $\lambda$ sampled from a uniform distribution over $[0,1]$ and the supervisory label of each image $y \in \{0,1\}^N$.
\begin{equation}
\tilde{y} = \lambda y_\mathrm{A} + (1-\lambda) y_\mathrm{B}.
\label{eq:img_mix_label}
\end{equation}

In this section, we directly use the mask $M$ in \cref{eq:img_mix_synth} as the provenance information.
That is, the provenance information corresponding to the supervisory labels $y_\mathrm{A}$ and $y_\mathrm{B}$ is defined as $\vI_\mathrm{A}=M$ and $ \vI_\mathrm{B}=\mathbf{1}-M$, respectively.
Therefore, a pixel $(u,v)$ in the synthetic image that contributes to the prediction of $y_\mathrm{A}$ is sampled from $x_\mathrm{A}$ and satisfies $\vI_\mathrm{A} (u,v)=1$.
The same applies to $y_\mathrm{B}$.
Through the above process, exact provenance information is obtained for each pixel in the synthetic image.

\subsubsection{Mixing Skeleton Sequences}\label{subsubsec:keypoint}
A synthetic learning method for mixing skeleton sequences~\citep{Hachiuma_2023_CVPR} combines pairs of skeleton sequences and supervisory labels, $(x_\mathrm{A}, y_\mathrm{A})$ and $(x_\mathrm{B}, y_\mathrm{B})$, together with the features $X_\mathrm{A}$ and $X_\mathrm{B}$ extracted from $x_\mathrm{A}$ and $x_\mathrm{B}$, respectively, following the same procedure as in the previous section.
Here, let $x \in \mathbb{R}^{P \times F \times K \times V}$ denote a skeleton sequence, where $P$ is the maximum number of skeletons detected in each frame, $F$ is the number of frames, $K$ is the number of joints, and $V$ is the input feature dimension for each joint (\eg, the detected position in the image).
These inputs are transformed by a DNN into per-skeleton features $X \in \mathbb{R}^{P \times F \times E}$, where $E$ is the feature dimension.
The features are then masked using the element-wise product $\odot$ as follows:
\begin{equation}
\hat{X}_\mathrm{A} = M \odot X_\mathrm{A},\qquad
\hat{X}_\mathrm{B} = (\mathbf{1}-M) \odot X_\mathrm{B},
\label{eq:skeleton_masking_features}
\end{equation}
where $M \in \{0,1\}^{P \times F \times E}$ is a binary spatio-temporal mask that replaces with 0 all elements in all frames for skeletons indexed from 1 to $P/T$ with 0 ($T$ is an integer, \eg, 2).
The synthesized feature $\tilde{X}$ is computed as follows.
\begin{equation}
\tilde{X} = \mathrm{MaxPool}(\hat{X}_A;\hat{X}_B).
\label{eq:maxpool}
\end{equation}
The supervisory label $\tilde{y}$ is computed in the same manner as in the previous section.
Note that, in \cref{eq:skeleton_masking_features}, an alternative synthesized scene can be obtained by swapping $M$ and $\mathbf{1}-M$.

As in the previous section, when the provenance information corresponding to supervisory labels $y_\mathrm{A}$ and $y_\mathrm{B}$ is defined as $\vI_\mathrm{A}=M$ and $ \vI_\mathrm{B}=\mathbf{1}-M$, respectively, each element indicates whether the corresponding feature originates from $X_\mathrm{A}$ or $X_\mathrm{B}$.
This provides exact provenance information in the spatio-temporal domain for each element of the synthesized feature $\tilde{X}$.

\subsubsection{Image Editing by Image Generation Models}\label{subsubsec:generative}
In prior work using image generation models~\citep{dunlap2023alia}, an edited image $\tilde{x} = G(x, p)$ is generated using a pretrained image generation model $G(\cdot)$, which takes an input image $x$ and a text prompt $p$ as inputs.
Here, $p$ is designed to modify regions of the image while excluding the target object.
Next, the transformed region is estimated by comparing the input image $x$ with the synthesized image $\tilde{x}$, and this information is used to compute the provenance.
A difference image $D \in \mathbb{R}^{H \times W}$ is computed, where the value at each pixel $(u,v)$ is defined as
\begin{align}
D(u,v) = \frac{1}{C}\sum_{c=1}^{C} \left| \tilde{x}_{uvc} - x_{uvc} \right|,
\end{align}
where $C$ is the number of channels.
Next, a threshold $\tau$ is obtained from $D$ using Otsu binarization~\citep{Otsu1979}, and the provenance information $\vI$ is computed as
\begin{align}
\vI(u,v) = M(u,v) =
\begin{cases}
0 & \text{if } D(u,v) > \tau,\\
1 & \text{otherwise}.
\end{cases}
\label{eq:alia_alpha}
\end{align}
In this paper, we assume that regions with $\vI(u,v)=0$ correspond to regions edited by the image generation model (\eg, background or co-occurring objects), whereas regions with $\vI(u,v)=1$ correspond to target regions that remain similar to the original image.
Therefore, $\vI$ functions as a mask that separates the target object from the background.

Based on the above, the proposed method can obtain annotation-free provenance information for a wide range of synthetic learning methods, including image/skeleton-sequence mixing methods and methods based on image generation models.

\subsection{Input Gradient Guidance}\label{subsec:provenanceguide}
The proposed method aims to encourage the model output for the target object to be attributable to the corresponding target regions in the input space.
To this end, we regularize the gradient of the model output $f_y(\tilde{x})$ (logit) for each class $y$ with respect to the input:
\begin{equation}
\nabla_{\tilde{x}} f_y(\tilde{x}) = \frac{\partial f_y(\tilde{x})}{\partial \tilde{x}}.
\end{equation}
The loss function $L_{\mathrm{PG}}$ is formulated based on this gradient; however, as described below, its computation differs depending on whether the supervisory label $\tilde{y}$ of the synthetic data is a soft label or a single hard label.

\subsubsection{Input Gradient Guidance for Soft Labels}
In image/skeleton-sequence mixing methods, $(\tilde{x}, \tilde{y})$ is constructed from $(x_\mathrm{A}, y_\mathrm{A})$ and $(x_\mathrm{B}, y_\mathrm{B})$, and the provenance information $\vI$ distinguishes elements originating from $x_\mathrm{A}$ ($\vI_\mathrm{A}(u,v)=1$) and elements originating from $x_\mathrm{B}$ ($\vI_\mathrm{B}(u,v)=1$).
In this setting, it is desirable that the logit for class $y_\mathrm{A}$, $f_\mathrm{A}(\cdot)$ depends only on the region from $x_\mathrm{A}$ ($\vI_\mathrm{A}(u,v)=M(u,v)=1$) and not on the region from $x_\mathrm{B}$ (\ie, where $\vI_\mathrm{B}(u,v)=1-M(u,v)=1$).
The same holds for class $y_\mathrm{B}$.

To enforce this, we define the provenance loss, which suppresses the occurrence of model input gradients for both classes in mutually irrelevant regions, as follows:
\begin{equation}
L_{\mathrm{PG}} = \left\| (\textbf{1}-M) \odot \nabla_{\tilde{x}} f_\mathrm{A}(\tilde{x}) + M \odot \nabla_{\tilde{x}} f_\mathrm{B}(\tilde{x}) \right\|_2^2.
\label{eq:ggloss}
\end{equation}
This loss suppresses both the case where the input gradients of $f_\mathrm{A}$ appear in regions originating from $x_\mathrm{B}$ and the reverse case, while encouraging each class to make predictions based only on the regions from which it originates.

\subsubsection{Input Gradient Guidance for Hard Labels}
In methods using image generation models, $\tilde{x}$ is synthesized but retains the original single hard label $y$.
If $\vI(u,v)=0$ denotes an edited region and $\vI(u,v)=1$ denotes an unedited target region, then ideally the logit for the supervisory label $y$, $f_{y}(\cdot)$, should depend only on the target region ($\vI(u,v)=M(u,v)=1$).
Therefore, we introduce the following loss function to suppress the occurrence of the model input gradients for $f_y$ in the edited region (1-$\vI(u,v)=1-M(u,v)=1$):
\begin{equation}
\label{eq:ggloss2}
L_{\mathrm{PG}} = \left\| (\mathbf{1}-M) \odot \nabla_{\tilde{x}} f_{y}(\tilde{x}) \right\|_2^2.
\end{equation}
This loss encourages the model to compute the logit by relying only on the target regions.

In both the soft-label and hard-label settings described above, $L_{\mathrm{PG}}$ acts as a regularization term that constrains the model input gradients based on the provenance information.

\subsection{Training Procedure}\label{subsec:training}
At each training iteration, mixing methods for image~\citep{Yun_2019_ICCV} and skeleton sequence~\citep{Hachiuma_2023_CVPR} randomly select training samples and regions to be mixed, and construct $(\tilde{x}, \tilde{y}, \vI)$.
For methods using image generation models, $(\tilde{x}, \vI)$ are prepared in advance as a synthetic dataset and used during training.
During model optimization, $L_{\mathrm{PG}}$ is computed using the provenance information $\vI$ according to \cref{eq:ggloss,eq:ggloss2}, and the model parameters are updated using the gradient of $L_{\mathrm{total}}$.

\section{Experiments}
\subsection{Datasets}
The effectiveness of the proposed method is evaluated for each synthetic learning method across multiple tasks. 
Following the experimental settings of prior work, image and skeleton-sequence mixing methods are evaluated on weakly supervised object localization and weakly supervised action detection, respectively, and methods using image generation models are evaluated on image classification.

\paragraph{CUB}
CUB-200-2011 (CUB)~\citep{wah2011caltech} is a fine-grained image classification dataset consisting of 200 bird classes, with each image annotated with a bounding box (BBox) for the target object and a class-level supervisory label. With relatively few training samples per class, it was proposed as a challenging dataset for image classification. In this study, CUB is used to evaluate both weakly supervised object localization and image classification.

\paragraph{iWildCam}
iWildCam~\citep{koh2021wilds} is an image classification dataset composed of images of wild animals from around the world, annotated with species labels. Because the capture environment, animal species, and imaging devices vary substantially across images, there exists a significant domain shift between the training and evaluation subsets.

\paragraph{Waterbirds}
Waterbirds~\citep{Sagawa2019Arxiv} is a two-class bird image classification dataset consisting of landbirds and waterbirds, in which the bird class and background environment are constructed to be strongly correlated, making it easy for models to make predictions based on the background rather than the foreground. It is therefore widely used as a benchmark for evaluating model robustness to spurious correlations.

\paragraph{UCF101-24}
UCF101-24 is a subset of UCF101~\citep{Soomro2012Arxiv} focusing on 24 action classes, with each video annotated with both action labels and per-frame BBoxes for the target person. Following prior work~\citep{Hachiuma_2023_CVPR}, this dataset is used to evaluate weakly supervised spatio-temporal action localization with skeleton sequences as input.

\begin{figure*}[t]
\centering
\begin{minipage}[t]{0.3\textwidth}
  \centering
  \includegraphics[width=0.8\linewidth]{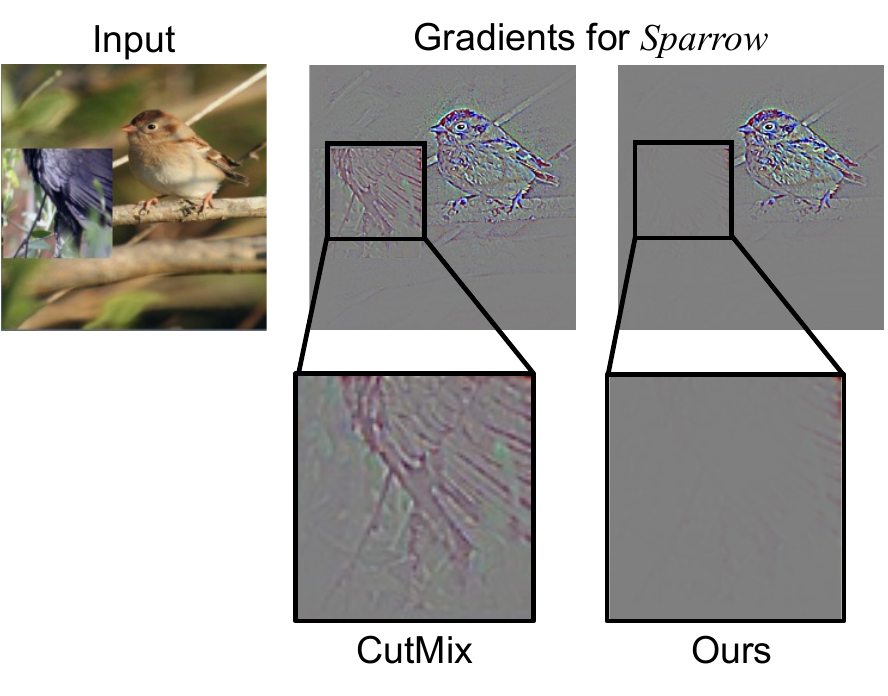}
  \captionof{figure}{Visualization of Guided Grad-class activation maps for each method on CutMix-synthesized images.}
  \label{fig:grad_results}
\end{minipage}
\hfill
\begin{minipage}[t]{0.28\textwidth}
  \centering
  \includegraphics[width=\linewidth]{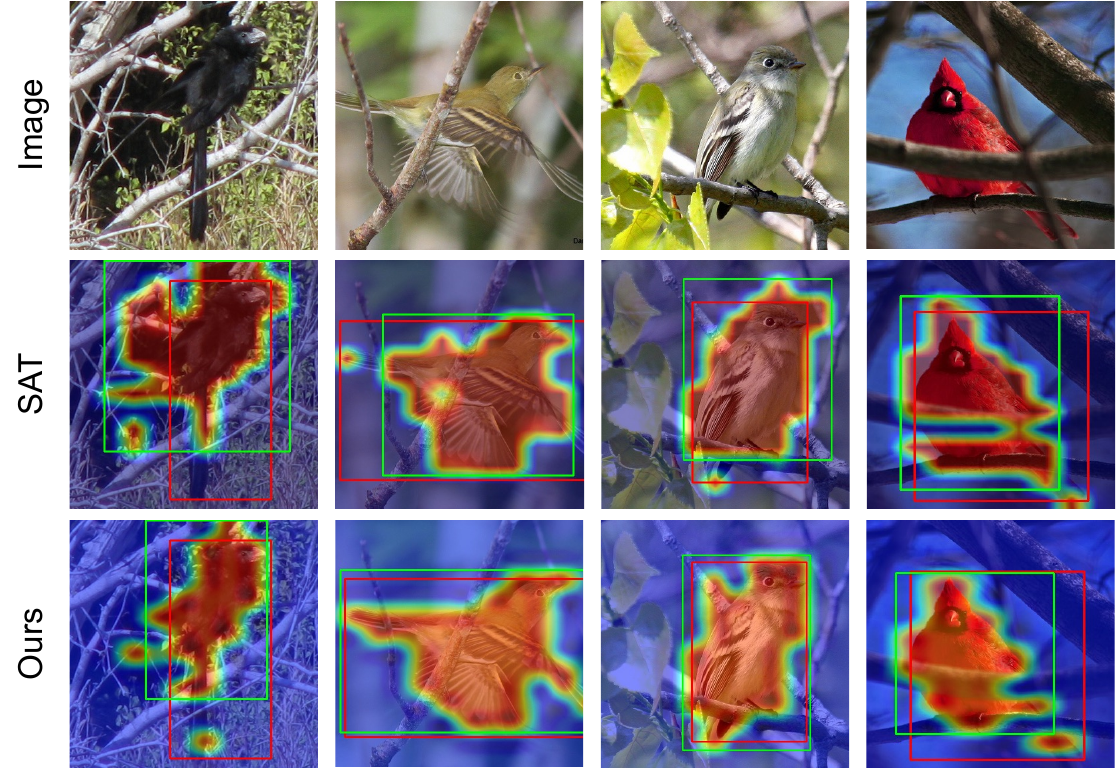}
  \captionof{figure}{Visualization of ground-truth BBoxes (\textcolor{red}{red}) and predictions of each method (\textcolor{green}{green}) on the CUB dataset.}
  \label{fig:experiment_visual}
\end{minipage}
\hfill
\begin{minipage}[t]{0.35\textwidth}
\centering
\vspace{-24mm}
\includegraphics[width=\linewidth]{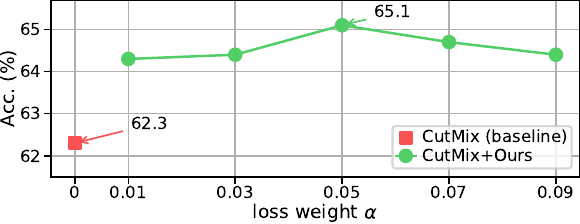}
\small
\captionof{figure}{Weakly supervised object localization accuracy as the coefficient $\alpha$ of the provenance loss in the total loss is varied on the CUB dataset.}
\label{fig:alpha_sensitivity}

\end{minipage}
\end{figure*}

\subsection{Experimental Settings}\label{subsec:exp_setting}

\subsubsection{Weakly Supervised Object Localization}
This task trains the model using only per-image supervisory labels, while at inference time predicting the object’s BBox and class. Two models were evaluated on the CUB dataset: VGG16~\citep{Simonyan_2015_ICLR}, pretrained on the ImageNet dataset~\citep{ImageNet}, and Spatial-Aware Token (SAT)~\citep{Wu_2023_ICCV}, which is a Transformer-based SoTA method. During training, data augmentation was applied using synthetic learning methods, including CutMix, and the proposed loss function was incorporated into each model. At inference time, BBoxes were estimated based on class activation maps (CAM)~\citep{Zhou_2016_CVPR} and Attention~\citep{Wu_2023_ICCV} obtained from each model. Evaluation used the implementation of Choe et al.\footnote{\url{https://github.com/clovaai/wsolevaluation}} with MaxBoxAccV2~\citep{Choe_2020_CVPR} as the metric; following prior work, adopted $\delta \in \{0.3, 0.5, 0.7\}$ as the IoU thresholds.

\subsubsection{Weakly Supervised Spatio-Temporal Action Localization}
This task trains the model using only video-level action labels, while at inference time detecting each person and predicting their actions frame by frame. The proposed method was incorporated into Structured Keypoint Pooling (SKP)~\citep{Hachiuma_2023_CVPR}, pretrained on Kinetics-400~\citep{Carreira2017CVPR}, and evaluated on UCF101-24. Following prior work~\citep{Hachiuma_2023_CVPR}, human skeletons were detected by HRNet~\citep{Sun_2019_CVPR}, and the skeletons of the top two persons with the highest joint detection scores in each frame were used as input. 
Data augmentation was then applied by mixing skeleton sequences across videos (hereafter referred to as BatchMix), and the proposed loss function was incorporated. Average Precision (AP) at an IoU threshold of 0.5 was used as the evaluation metric.

\subsubsection{Image Classification}
CUB, iWildCam, and Waterbirds assess fine-grained classification accuracy, robustness to domain shift, and robustness to background spurious correlations, respectively, and were used in this experiment to evaluate the proposed method from different perspectives. Following the experimental setting of prior work using image generation models~\citep{dunlap2023alia}, input images were edited with an image generation model before the proposed method was introduced. ResNet-50~\citep{He_2016_CVPR}, pretrained on the ImageNet dataset, was used as the model. Classification accuracy (Top-1) was used as the evaluation metric. 

All experiments were conducted on two NVIDIA GPUs (Quadro GV100 and Quadro P6000).

\begin{table}[t]
\centering
\caption{Comparison of weakly supervised object localization accuracy between baseline methods and the proposed method on the CUB dataset. Underlined values are reference values whose trends are inconsistent across IoU thresholds.
}
\label{table:vgg16_sat_cub}
   \scalebox{0.7}{
\begin{tabular}{l|l|cccc}
\hline 
\multirow{2}{*}{Backbone} & \multirow{2}{*}{Method} &  \multicolumn{4}{c}{MaxBoxAccV2 (\%)} \\
 & & $\delta$ = 0.3 & 0.5 & 0.7 & Mean \\
\hline 
\multirow{6}{*}{VGG16~\citep{Simonyan_2015_ICLR}} & CAM~+CutMix~\citep{Yun_2019_ICCV} &  91.1 & 67.3&  \underline{\textcolor{black!30}{28.6}} & 62.3\\
& ~~~~~~~~~~+Ours & \textbf{96.8} &  \textbf{74.6} & \textbf{23.1} & \textbf{65.1} \\
\cline{2-6}
& CAM~+ResizeMix~\citep{qin2020resizemix} &  92.4 & 62.4& 18.1 & 57.6\\
& ~~~~~~~~~~+Ours & \textbf{95.9} &  \textbf{70.6} & \textbf{20.1} & \textbf{62.2} \\
\cline{2-6}
& CAM~+PuzzleMix~\citep{kim2020puzzle} &  94.6 & 61.3& 14.4 & 56.8\\
& ~~~~~~~~~~+Ours & \textbf{95.2} &  \textbf{64.8} & \textbf{14.6} & \textbf{58.2} \\
\hline

\multirow{6}{*}{DeiT-S~\citep{Touvron_2021_ICML}} &TS-CAM~\citep{Gao_2021_ICCV} &   98.9 & 87.7 & 49.9 & 78.8 \\
& SCM~\citep{Bai_2022_CVPR} & 99.6 & 96.6 & 71.7 & 89.3 \\ 
& GTFormer~\citep{yang2024pro2sam} & - & 97.4 & 77.0 & - \\ 
& SAT~\citep{Wu_2023_ICCV} & 99.8 & 97.4 & 76.9 & 91.4 \\
& ~~+CutMix~\citep{Yun_2019_ICCV} & 99.8 & 97.2 & 77.4 & 91.5 \\
& ~~~~+Ours & \textbf{99.9} & \textbf{97.5} & \textbf{78.8} & \textbf{92.1} \\
\hline
\end{tabular}
}
\end{table}

\begin{table*}[t]
\centering
\small

\begin{minipage}[t]{0.32\textwidth}
    \centering
    \captionof{table}{Comparison of weakly supervised spatio-temporal action localization accuracy between baseline methods and the proposed method on the UCF101-24 dataset.}
    \label{tab:act_loc}
    \scalebox{0.7}{
    \begin{tabular}{l|c|c}
    \hline
    Method & Input & AP \\ \hline
    Ch\'{e}ron~\etal~\citep{Cheron2018Neurips} & \multirow{2}{*}{RGB} & 17.7 \\ 
    Anurag~\etal~\citep{Anurag2020ECCV} &  & 35.0 \\ \hline
    SKP~\citep{Hachiuma_2023_CVPR} & \multirow{3}{*}{Skeleton} & 37.4 \\
    ~~+BatchMix & & 38.0 \\
    ~~~~+Ours &  & \textbf{39.7} \\ \hline
    \end{tabular}
    }
\end{minipage}
\hfill
\begin{minipage}[t]{0.32\linewidth}
\vspace{0pt}
\centering
\small
\captionof{table}{Comparison of image classification accuracy between baseline methods and the proposed method.}
\scalebox{0.7}{
\begin{tabular}{l|c|c|c}
\hline 
Method & CUB & iWildCam & Waterbirds \\
\hline 
Baseline & 70.8 & 75.0 & 62.2 \\
Random Augment & 67.8 & 71.3 & 64.0 \\
CutMix~\citep{Yun_2019_ICCV} & 68.0 & 77.2 & 63.4 \\
\hline
ALIA~\citep{dunlap2023alia} & 71.7 & 83.5 & 71.4 \\
~~+Ours (Mean) & 72.0$_{\pm0.1}$ & 84.4$_{\pm0.7}$ & 80.7$_{\pm1.6}$ \\
~~+Ours (Max) & \textbf{72.1} & \textbf{85.1} & \textbf{82.3} \\
\hline
\end{tabular}}
\label{tab:cls}
\end{minipage}
\hfill
\begin{minipage}[t]{0.31\linewidth}
\vspace{0pt}
\centering
\small
\captionof{table}{Comparison of hyperparameter tuning efficiency for each method in weakly supervised object localization on the CUB dataset. VGG16 is used as the base model.}
\scalebox{0.7}{
\begin{tabular}{l|c|c|c}
\hline
Method & \#Runs (search space) & Acc. (\%) & Total (h) \\
\hline
CutMix & 16 ($lr$, $wd$) & 62.3 & 31 \\ \hline
\multirow{2}{*}{~~+Ours} & 18 ($lr$, $wd$, $\alpha$) & 64.7 & 11 \\
                         & 48 ($lr$, $wd$, $\alpha$) & 65.1 & 30 \\
\hline
\end{tabular}}
\label{tab:grid_search}
\end{minipage}

\end{table*}

\begin{table*}[t]
\centering
\small

\begin{minipage}[t]{0.39\linewidth}
\vspace{0pt}
\centering
\small
\captionof{table}{Comparison of training efficiency for each method in weakly supervised object localization on the CUB dataset. VGG16 is used as the base model.}
\scalebox{0.7}{
\begin{tabular}{l|c|c|c|c|c|c}
\hline
Method & BS & Acc. (\%) & Epochs & Sec/Epoch & Total (h) & Mem. \\
\hline
CutMix & 32 & 62.3 & 50 & 140 & 1.9 & 10GB \\ \hline
\multirow{2}{*}{~~+Ours} & 32 & 65.1 & 15 & 150 & 0.6 & 14GB \\
                         & 16 & 64.2 & 15 & 180 & 0.8 & 7GB \\
\hline
\end{tabular}}
\label{tab:cost_eff}
\end{minipage}
\hfill
\begin{minipage}[t]{0.2\textwidth}
    \centering
    \captionof{table}{Comparison of the proposed method across mask-image patterns of provenance information. See Sec.~\ref{sec:mask} for details.}
    \label{tab:r1_major4_controls}
    \scalebox{0.7}{
    \begin{tabular}{c|c}
    \hline
    Mask & Acc. \\
    \hline
    Random & 60.5 \\
    Unmasked & 61.1 \\
    Ours & \textbf{65.1} \\
    \hline
    \end{tabular}
    }
\end{minipage}
\hfill
\begin{minipage}[t]{0.36\textwidth}
    \centering
    \captionof{table}{Comparison of the proposed method with respect to the quality of the difference-mask image before and after image editing used as provenance information. See Sec.~\ref{sec:mask} for details.}
    \label{tab:mask_noise_sens}
    \scalebox{0.7}{
    \begin{tabular}{l|c|c|c}
    \hline
    Mask & $\Delta$FG area & CUB & Waterbirds \\
    \hline
    Ours & 0 & 72.1 & 81.8 \\
    \hline
    \multirow{2}{*}{+Dilation} & +10\% & 72.0 & 81.5 \\
                               & +30\% & 71.8 & 81.2 \\
    \hline
    \multirow{2}{*}{+Erosion}  & -10\% & 71.9 & 81.7 \\
                               & -30\% & 71.5 & 81.4 \\
    \hline
    \end{tabular}
    }
\end{minipage}

\end{table*}

\subsection{Comparative Experiments Against Baseline Methods}

\subsubsection{Weakly Supervised Object Localization}\label{subsec:cutmix}
\cref{table:vgg16_sat_cub} shows the localization accuracy of baseline methods and the proposed method on the CUB dataset, where the proposed method is introduced into synthetic learning methods, including CutMix. Among VGG16-based methods, incorporating the proposed method into CAM achieves the best mean accuracy of $65.1\%$.
Even when the proposed method is introduced into other synthetic learning methods, namely ResizeMix~\citep{qin2020resizemix} and PuzzleMix~\citep{kim2020puzzle}, accuracy improves by 5.1 pp and 1.3 pp, respectively.
With DeiT-S as the base model, the SoTA method SAT achieves a mean accuracy of 91.4\%; fine-tuning with CutMix slightly improves this to 91.5\%, and incorporating the proposed method further improves it to 92.1\%.
These results confirm that provenance-information-based input gradient guidance is broadly effective for image-mixing synthetic learning methods.

\cref{fig:grad_results} shows Guided Grad-CAM~\citep{Selvaraju_2020_IJCV} visualization results for the target object (Sparrow) on CutMix-synthesized images. Without the proposed method, gradients appear in regions unrelated to the target object, indicating that model predictions depend on synthesized regions; with input gradient guidance, gradients are suppressed over synthesized regions and concentrated on the true target-object regions.
\cref{fig:experiment_visual} shows the attention heatmap visualization results for evaluation samples using SAT as the base model. The attention distributions more closely match the bird silhouette, indicating that input gradient guidance prevents SAT from attending to background regions and suppresses spurious correlations. Furthermore, the proposed method not only reduces spurious correlations but also enables more accurate segmentation of the target object.

\subsubsection{Weakly Supervised Spatio-Temporal Action Localization}\label{subsec:batchmix}
\cref{tab:act_loc} shows the localization accuracy of baseline methods and the proposed method on the UCF101-24 dataset. Introducing BatchMix into SKP~\citep{Hachiuma_2023_CVPR} (baseline: 37.4\%) demonstrates the effect of skeleton-sequence mixing augmentation, improving AP to 38.0\%. Incorporating the proposed method further improves AP by 1.7 pp (39.7\%). These gains demonstrate that leveraging provenance information from skeleton-sequence mixing to perform input gradient guidance encourages the model to attend to action-relevant skeletons in the spatio-temporal domain, thereby improving localization accuracy.

\subsubsection{Image Classification}\label{subsec:alia}
\cref{tab:cls} shows image classification accuracy on CUB, iWildCam, and Waterbirds, where the proposed method is introduced into ALIA~\citep{dunlap2023alia}, which edits training samples using an image generation model.

On CUB, the baseline accuracy without ALIA is 70.8\%; introducing ALIA improves it to 71.7\%, and further incorporating the proposed method consistently improves it to 72.0\%. 
On Waterbirds, where robustness to spurious correlations is required, the substantial gain brought by introducing the proposed method to ALIA (9.6 pp) directly supports the effectiveness of input gradient guidance for suppressing spurious correlations.

Taken together, these results confirm that the proposed method’s provenance-guided input gradient regularization works as intended across multiple tasks and modalities. 

\subsection{Ablation Study}\label{subsec:ablation}

\subsubsection{Effect of Introducing the Provenance Loss}\label{sec:alpha}
\cref{fig:alpha_sensitivity} shows the results obtained by varying the contribution of the provenance loss $L_{\mathrm{PG}}$ in the total loss using the coefficient $\alpha$ in \cref{eq:overview_total}.
When $\alpha$ is varied during training for weakly supervised object localization on the CUB dataset, the localization accuracy changes smoothly, indicating that the learning process does not strongly depend on a specific value of $\alpha$.
Furthermore, the accuracy achieved with the introduction of the provenance loss consistently exceeds that of CutMix, demonstrating that stable performance can be obtained over the range $\alpha\in[0.01,0.09]$.

\subsubsection{Evaluation of Training Efficiency}
\cref{tab:cost_eff} compares training efficiency before and after introducing the proposed method into the baseline CutMix for weakly supervised object localization on the CUB dataset. When the batch size is fixed at 32, the proposed method requires higher memory usage and longer training time per epoch owing to the second-order loss differentiation involved in gradient guidance. However, this overhead can be mitigated by reducing the batch size (\eg, from 32 to 16), with no resulting barrier in terms of localization accuracy, convergence time, or memory usage.
In addition, second-order differentiation of the loss was computed using PyTorch autograd\footnote{\url{https://pytorch.org/docs/stable/autograd.html}} and AMP\footnote{An abbreviation for PyTorch’s Automatic Mixed Precision technique for efficient execution of DNN models.}, while the loss function was computed in FP32 to avoid numerical instability during training, thereby preventing NaN/Inf losses.

\subsubsection{Evaluation of Hyperparameter Tuning Efficiency}
\cref{tab:grid_search} shows hyperparameter tuning results before and after introducing the proposed method into the baseline CutMix for weakly supervised object localization on the CUB dataset. The hyperparameters requiring tuning for CutMix are the learning rate ({\it lr}) and weight decay ({\it wd}); introducing the proposed method additionally requires tuning the coefficient $\alpha$ described in Sec.~\ref{sec:alpha}.
When the number of tuning trials (\#Runs) is comparable before and after introducing the proposed method (16 vs.\ 18) and when comparable tuning time is spent (31 vs.\ 30), the proposed method improves accuracy over CutMix in both cases, confirming that introducing the proposed method does not degrade hyperparameter tuning efficiency.

\subsubsection{Evaluating the Effectiveness of Input Gradient Guidance Using Provenance Information}\label{sec:mask}
To verify the effectiveness of provenance-information-based input gradient guidance, the provenance mask is replaced with alternative patterns and the resulting accuracy is compared. \cref{tab:r1_major4_controls} shows object localization accuracy on the CUB dataset under two settings: (1) random binary values assigned per pixel (Random), and (2) gradient regularization applied over the entire input image without a mask (Unmasked). Both settings perform worse than the proposed method, confirming that the accuracy improvement is attributable specifically to the use of provenance information, not to gradient regularization alone.

In addition, in the evaluation of image classification for image editing methods, \cref{tab:mask_noise_sens} compares classification accuracy after applying dilation and erosion to the target regions ($\vI(u,v)=1$ regions) in the difference-mask image before and after image editing described in Sec.~\ref{subsubsec:generative}.
Dilation and erosion of the target regions correspond to mistakenly including background or foreground pixels, respectively, in the target-object region. Provided that the ratio of the changed target region area (\ie, pixels mistakenly guided by input gradient guidance) remains within 30\%, the performance drop stays below 1\%, confirming that input gradient guidance is robust to moderate imprecision in the provenance mask.

\section{Conclusion}
This paper proposed a learning framework for synthetic data that uses provenance information in the input space obtained during the synthetic process as an auxiliary supervisory signal to promote the acquisition of representations focused on target regions. By introducing input gradient guidance that separates input gradients based on target and non-target regions and suppresses gradients over non-target regions, the model output is prevented from depending on non-target regions, such as background and synthetic artifacts, enabling the model to directly learn discriminative representations of target regions. Evaluations across multiple tasks and modalities, including weakly supervised object localization, weakly supervised spatio-temporal action localization, and image classification, confirmed the effectiveness and generality of the proposed method. 

{
    \small
    \bibliographystyle{ieeenat_fullname}
    \bibliography{main}
}

\clearpage
\setcounter{page}{1}

\twocolumn[{%
\renewcommand\twocolumn[1][]{#1}%
\maketitlesupplementary
\centering

\captionof{table}{Hyperparameters of each dataset during training.}
\label{tab:hp}
\vspace{2mm}

\scalebox{0.7}{
\begin{tabular}{c|cccccc} \hline 
    Training dataset 
& \cellcolor{blue!15} UCF101-24~\citep{Soomro2012Arxiv}  
& \multicolumn{3}{c}{\cellcolor{red!15} CUB~\citep{wah2011caltech}} 
& \cellcolor{green!15} iWildCam~\citep{koh2021wilds} 
& \cellcolor{yellow!20} Waterbirds~\citep{Sagawa2019Arxiv} \\ \hline
    Backbone & HRNet & VGG16 & DeiT-S & ResNet-50 & ResNet-50 & ResNet-50 \\
    Mixing probability  & 1.0 & 1.0 & 0.79 & - & - & - \\ 
    Amount of & \multirow{2}{*}{-} & \multirow{2}{*}{-} & \multirow{2}{*}{-} & \multirow{2}{*}{1000} & \multirow{2}{*}{2224} & \multirow{2}{*}{839}\\
    Augmented Data Added. & & & & & & \\
    Loss weight $\alpha$ in \cref{eq:overview_total} & 0.01 & 0.05 & 0.01 & 0.1 & 0.1 & 0.1 \\
    Optimizer        & SGD & SGD & AdamW & SGD & SGD & SGD \\
    Number of epochs & 40 & 15 & 10 & 20 & 10 & 20 \\ 
    Batch size       & 30 & 32 & 32 & 128 & 128 & 128 \\ 
    Learning rate    & 7.5e-3 & 1e-2 & 1e-5 & 1e-3 & 1e-3 & 1e-3 \\ 
    LR scheduler     & linear & linear & linear & cosine & cosine & cosine \\  
    Weight decay     & 2.5e-5 & 5e-4 & 5e-3 & 1e-5 & 1e-4 & 1e-4 \\ 
    Momentum         & \multicolumn{6}{c}{0.9} \\ \hline
\end{tabular}
}

\vspace{1em}
}]

\section{Implementation Details}
In this section, we provide details on data augmentation and training hyperparameters.

As described in \cref{subsubsec:generative}, provenance information is derived by computing a difference image between the generated and source images, followed by Otsu binarization~\citep{Otsu1979}, to produce a binary mask distinguishing target regions from non-target regions. 

\begin{figure*}[tbp]
    \centering
    \begin{minipage}{0.48\textwidth}
        \centering
        \includegraphics[width=\linewidth,height=\textheight,keepaspectratio]{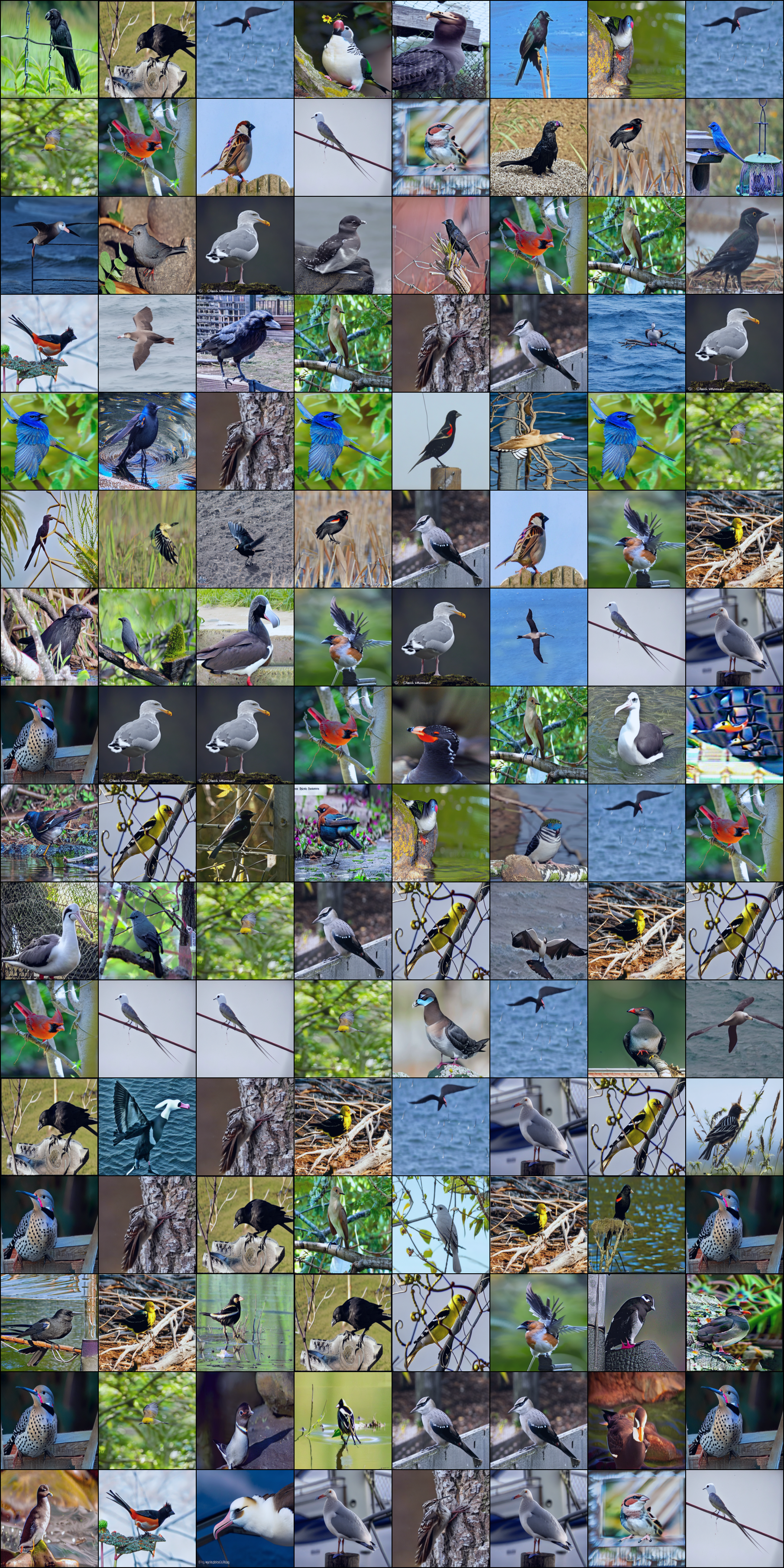}
        \caption*{(a) Generated images from image editing synthesis.}
    \end{minipage}
    \hfill
    \begin{minipage}{0.48\textwidth}
        \centering
        \includegraphics[width=\linewidth,height=\textheight,keepaspectratio]{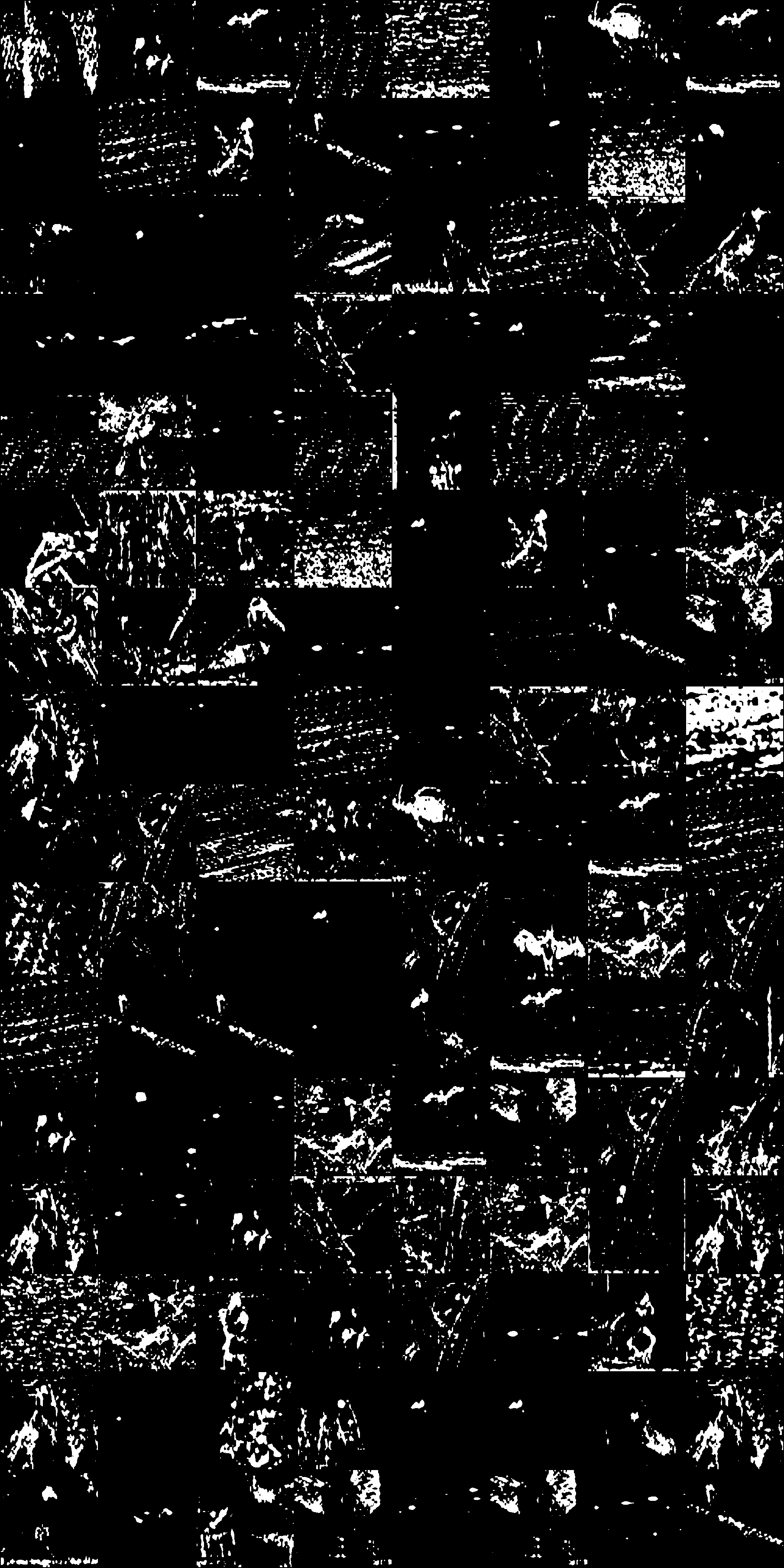}
        \caption*{(b) Provenance masks derived from difference images.}
    \end{minipage}

    \caption{Visualization of image editing synthesis and the corresponding provenance masks.}
    \label{fig:visualize_generating_based_edit}
\end{figure*}

As illustrated in \cref{fig:visualize_generating_based_edit}, paired examples of synthetic-data samples alongside their corresponding provenance masks are visualized. Only a subset of these difference-based masks accurately captures the true target regions; many include residual background or synthetic artifacts. Improving provenance information quality, for example by leveraging cross-attention signals from the image generation model instead of relying solely on difference images, is an important direction for future work.

\subsection{Hyperparameters}
The hyperparameters for each dataset and synthesis setting are summarized in \cref{tab:hp}. Two types of synthesis are considered: mixing-based synthetic learning methods for localization and image editing methods for classification.

For the mixing-based synthetic learning methods (BatchMix and CutMix in \cref{subsec:batchmix,subsec:cutmix}), skeleton inputs with an HRNet backbone are used for weakly supervised spatio-temporal action localization on UCF101-24, and image inputs with VGG16 and DeiT-S backbones are used for weakly supervised object localization on CUB. The mixing probability (second row of \cref{tab:hp}) is set to $1.0$ for UCF101-24, and to $1.0$ and $0.79$ for CUB with VGG16 and DeiT-S backbones, respectively.

For image editing methods (ALIA in \cref{subsec:alia}) on CUB, iWildCam, and Waterbirds, we use ResNet-50 as the backbone for image classification.
Instead of using a mixing probability, we control the number of generated images added to each training set (``Amount of Augmented Data Added’’ in \cref{tab:hp}). Following ALIA~\citep{dunlap2023alia}, we generate $1000$ additional images for CUB, $2224$ for iWildCam, and $839$ for Waterbirds.

The loss balancing weight $\alpha$ in \cref{eq:overview_total} is selected for each dataset and task (see \cref{tab:hp}) and kept fixed during training.
For skeleton-based localization on UCF101-24, we train HRNet with SGD for $40$ epochs using a linear schedule.
For weakly supervised object localization on CUB, we train VGG16 for $15$ epochs with SGD and DeiT-S for $10$ epochs with AdamW, both using a linear schedule.
For experiments with the image editing method on CUB, iWildCam, and Waterbirds, we train ResNet-50 with SGD for $20$, $10$, and $20$ epochs, respectively, using a cosine schedule.
The batch size, learning rate, weight decay, and optimizer are provided in \cref{tab:hp}.

Across all settings, we use a momentum of $0.9$.
The learning rate, weight decay, and $\alpha$ are tuned via coarse-to-fine grid search, while other hyperparameters follow standard practice for each backbone and dataset.

\section{Ablation Study}

\subsection{Training Efficiency}
As accuracy results are presented in \cref{subsec:cutmix,subsec:alia}, this section focuses on training efficiency.
We measure efficiency by the number of epochs required to reach peak validation performance (``Best epoch'') under identical setups in \cref{subsec:exp_setting}, as summarized in \cref{tab:skg_cutmix_backbone,tab:cub_iwildcam_alia_skg}.

\subsubsection{Image mixing}
Compared with CutMix, our method reduces the Best epoch from $50 \to 15$ on VGG16 ($\approx 3.3\times$ fewer epochs) and from $30 \to 10$ on DeiT-S ($3\times$ fewer).
This consistent reduction indicates faster and more stable optimization across both CNN and transformer backbones.
\subsubsection{Image Editing by Image Generation Models}
On CUB, our method reaches peak performance in $10$ epochs, compared with $15$ for ALIA ($1.5\times$ faster).
On iWildCam, it converges in $5$ epochs, compared with $10$ for ALIA ($2\times$ faster).
On Waterbirds, it reaches peak performance in $10$ epochs, compared with $15$ for ALIA ($1.5\times$ faster).
These consistent trends across datasets with different distribution shifts suggest that provenance-guided regularization improves sample efficiency.

\begin{table*}[t]
\centering
\begin{minipage}[t]{0.38\textwidth}
\centering
\small
\captionof{table}{ Accuracy and training efficiency comparison of mix-based image synthesis (WSOL on CUB).}
\scalebox{0.9}{
\begin{tabular}{l|c|cc}
\hline
\multirow{2}{*}{Method} & \multirow{2}{*}{Backbone} & \multirow{2}{*}{Acc.~(\%) $\uparrow$} & Best \\
 & & & epoch $\downarrow$ \\
\hline
CutMix & \multirow{2}{*}{VGG16~\citep{Simonyan_2015_ICLR}} & 62.3 & 50 \\
~~+Ours & & 65.1 & \textbf{15} \\ \hline
CutMix & \multirow{2}{*}{DeiT-S~\citep{Touvron_2021_ICML}} & 91.5 & 30 \\
~~+Ours & & 92.0 & \textbf{10} \\
\hline
\end{tabular}
\label{tab:skg_cutmix_backbone}
\hspace{0.1mm}
}
\end{minipage}
\hfill
\begin{minipage}[t]{0.61\textwidth}
\centering
\small
\captionof{table}{ Accuracy and training efficiency comparison on CUB, iWildCam, and Waterbirds for the image editing method.}
\scalebox{0.85}{
\begin{tabular}{l|cc|cc|cc}
\hline
\multirow{2}{*}{Method} 
& \multicolumn{2}{c}{CUB} 
& \multicolumn{2}{c}{iWildCam} 
& \multicolumn{2}{c}{Waterbirds} \\
& Acc.~(\%) $\uparrow$ & Best epoch $\downarrow$ 
& Acc.~(\%) $\uparrow$ & Best epoch $\downarrow$
& Acc.~(\%) $\uparrow$ & Best epoch $\downarrow$ \\
\hline
ALIA & 71.7 & 15 & 83.5 & 10 & 71.4 & 15 \\
~~+Ours & 72.0 & \textbf{10} & 84.4 & \textbf{5} & 80.7 & \textbf{10} \\
\hline
\end{tabular}
\label{tab:cub_iwildcam_alia_skg}
}
\end{minipage}

\end{table*}

\clearpage

\end{document}